\documentclass[11pt,letterpaper]{article}

\usepackage[letterpaper]{geometry}
\usepackage{acl2012}
\usepackage{times}
\usepackage{latexsym}
\usepackage{amsmath}
\usepackage{multirow}
\usepackage{url}
\usepackage{xcolor}
\usepackage{amsfonts}
\usepackage{bm}
\usepackage[sort]{natbib}
\usepackage{enumitem}
\usepackage{subcaption}
\usepackage{graphicx}

\setlist[description]{leftmargin=-2em,style=unboxed}

\makeatletter
\newcommand{\@BIBLABEL}{\@emptybiblabel}
\newcommand{\@emptybiblabel}[1]{}
\makeatother
\usepackage[hidelinks]{hyperref}

\title{Learning Structured Text Representations}

\author{Yang Liu \and Mirella Lapata \\
  Institute for Language, Cognition and Computation
  \\
  School of Informatics, University of Edinburgh
  \\ 
  10 Crichton Street, Edinburgh EH8 9AB \\
  {\tt yang.liu2@ed.ac.uk,mlap@inf.ed.ac.uk}
}

\date{}

\begin{document}
\maketitle
\begin{abstract}
  In this paper, we focus on learning structure-aware document
  representations from data without recourse to a discourse parser or
  additional annotations. Drawing inspiration from recent efforts to
  empower neural networks with a structural bias
  \citep{kim2017structured,cheng2016long}, we propose a model that can
  encode a document while automatically inducing rich structural
  dependencies. Specifically, we embed a differentiable non-projective
  parsing algorithm into a neural model and use attention mechanisms
  to incorporate the structural biases. Experimental evaluations across
  different tasks and datasets show that the proposed model achieves
  state-of-the-art results on document modeling tasks while inducing
  intermediate structures which are both interpretable and meaningful.

%
%
\end{abstract}

\section{Introduction}
Document modeling is a fundamental task in Natural Language Processing
useful to various downstream applications including topic
labeling~\citep{DBLP:conf/uai/XieX13},
summarization~\citep{chen2016distraction,Wolf:Gibson:2006}, sentiment
analysis~\citep{bhatia2015better}, question answering
\citep{Verberne:ea:2007}, and machine translation
\citep{meyer-webber:2013:DiscoMT}.


Recent work provides strong evidence that better document
representations can be obtained by incorporating structural
knowledge~\citep{ji2017neural,bhatia2015better,yang2016hierarchical}.
Inspired by existing theories of discourse, representations of
document structure have assumed several guises in the literature, such
as trees in the style of Rhetorical Structure Theory
\citep[RST;][]{mann1988rhetorical}, graphs
\citep{lin-ng-kan:2011:ACL-HLT2011,Wolf:Gibson:2006}, entity
transitions \citep{Barzilay:Lapata:08}, or combinations thereof
\citep{lin-ng-kan:2011:ACL-HLT2011,mesgar-strube:2015:*SEM2015}.  The
availability of discourse annotated corpora
\citep{carlson2003building,prasad2008penn} has led to the development
of off-the-shelf discourse parsers
\citep[e.g.,][]{feng-hirst:2012:ACL2012,liu-lapata:2017:EMNLP2017}, and
the common use of trees as representations of document structure. For
example, \cite{bhatia2015better} improve document-level sentiment
analysis by reweighing discourse units based on the depth of RST
trees, whereas \cite{ji2017neural} show that a recursive neural
network built on the output of an RST parser benefits text
categorization in learning representations that focus on salient
content.

Linguistically motivated representations of document structure rely on
the availability of annotated corpora as well as a wider range of
standard NLP tools (e.g., tokenizers, pos-taggers, syntactic
parsers). Unfortunately, the reliance on labeled data, which is both
difficult and highly expensive to produce, presents a major obstacle
to the widespread use of discourse structure for document modeling.  Moreover, despite recent
advances in discourse processing, the use of an external parser often
leads to pipeline-style architectures where errors propagate to later
processing stages, affecting model performance.

It is therefore not surprising that there have been attempts to induce
document representations directly from data without recourse to a
discourse parser or additional annotations.  The main idea is to
obtain \emph{hierarchical} representations by first building
representations of sentences, and then aggregating those into a
document representation \citep{tang2015document,tang2015learning}.
\cite{yang2016hierarchical} further demonstrate how to implicitly
inject structural knowledge onto the representation using an attention
mechanism \citep{bahdanau2014neural} which acknowledges that sentences
are differentially important in different contexts. Their model learns
to pay more or less attention to individual sentences when
constructing the representation of the document.

\begin{figure}[t]
  \centering
  \includegraphics[width=3.1in]{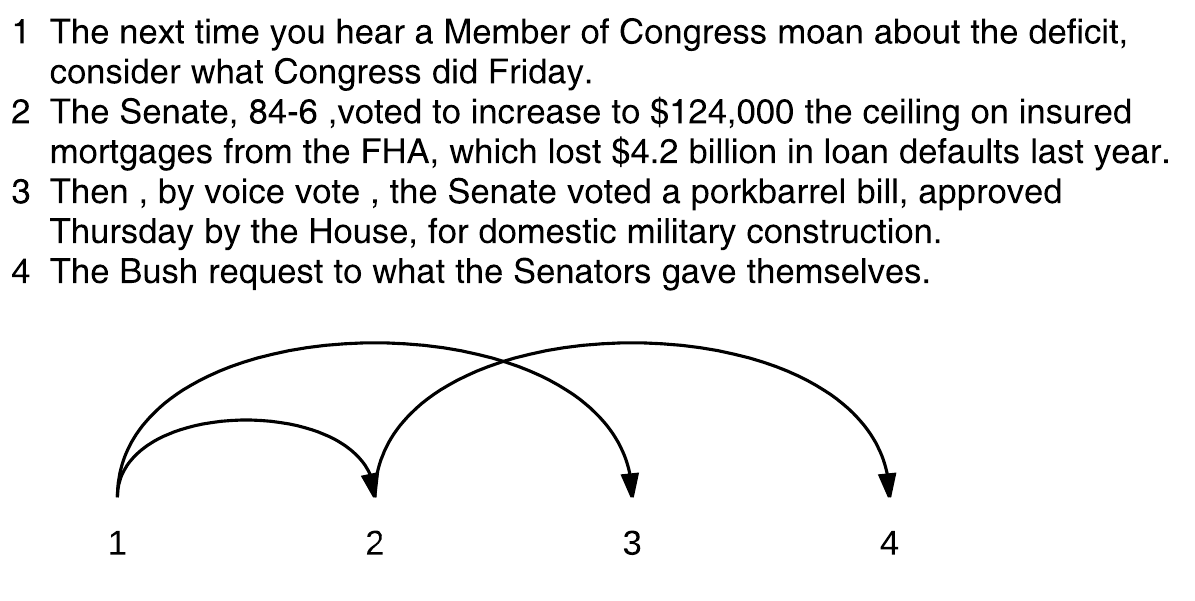}
  \caption{The document is analyzed in the style of Rhetorical
    Structure Theory~\protect\citep{mann1988rhetorical}, and
    represented as a dependency tree following the conversion algorithm
    of~\protect\cite{hayashi2016empirical}. }
  \label{fig:discourse}
\vspace{-2ex}
\end{figure}

Our work focus on learning deeper structure-aware document
representations, drawing inspiration from recent efforts to empower
neural networks with a structural bias
\citep{cheng2016long}. \cite{kim2017structured} introduce structured
attention networks which are generalizations of the basic attention
procedure, allowing to learn sentential representations while
attending to partial segmentations or subtrees. Specifically, they
take into account the dependency structure of a sentence by viewing
the attention mechanism as a graphical model over latent variables.
They first calculate unnormalized pairwise attention scores for all
tokens in a sentence and then use the inside-outside algorithm to
normalize the scores with the marginal probabilities of a dependency
tree.  Without recourse to an external parser, their model learns
meaningful task-specific dependency structures, achieving competitive
results in several sentence-level tasks. However, for document modeling, this
approach has two drawbacks. Firstly, it does not consider
non-projective dependency structures, which are common in
document-level discourse
analysis~\citep{lee2006complexity,hayashi2016empirical}.  
As
illustrated in Figure~\ref{fig:discourse}, the tree structure of a
document can be flexible and the dependency edges may cross.
Secondly, the inside-outside algorithm involves a dynamic programming
process which is difficult to parallelize, making it impractical for
modeling long documents.\footnote{In our experiments, adding the
  inside-outside pass increases training time by a factor of~10.}

In this paper, we propose a new model for representing documents while
automatically learning richer structural dependencies. Using a variant
of Kirchhoff's Matrix-Tree Theorem~\citep{tutte1984graph}, our model
implicitly considers non-projective dependency tree structures. We
keep each step of the learning process differentiable, so the model
can be trained in an end-to-end fashion and induce discourse
information that is helpful to specific tasks without an external
parser. The inside-outside model of \cite{kim2017structured} and our
model both have a~$O(n^3)$ worst case complexity. However, major
operations in our approach can be parallelized efficiently on GPU
computing hardware. Although our primary focus is on document modeling,
there is nothing inherent in our model that prevents its application
to individual sentences. Advantageously, it can induce non-projective
structures which are required for representing languages with free or
flexible word order \citep{mcdonald-satta:2007:IWPT2007}.

Our contributions in this work are threefold: a model for learning
document representations whilst taking structural information into
account; an efficient training procedure which allows to compute
document level representations of arbitrary length; and a large scale
evaluation study showing that the proposed model performs
competitively against strong baselines while inducing intermediate
structures which are both interpretable and meaningful.

\section{Background}
\label{sec:background}

In this section, we describe how previous work uses the attention
mechanism for representing individual sentences.  The key idea is to
capture the interaction between tokens within a sentence, generating a
context representation for each word with weak structural
information. This type of \emph{intra-sentence} attention encodes
relationships between words within each sentence and differs from
\emph{inter-sentence} attention which has been widely applied to
sequence transduction tasks like machine translation
\citep{bahdanau2014neural} and learns the latent alignment
\emph{between} source and target sequences.

\begin{figure}[t]
  \centering
  \includegraphics[width=3in]{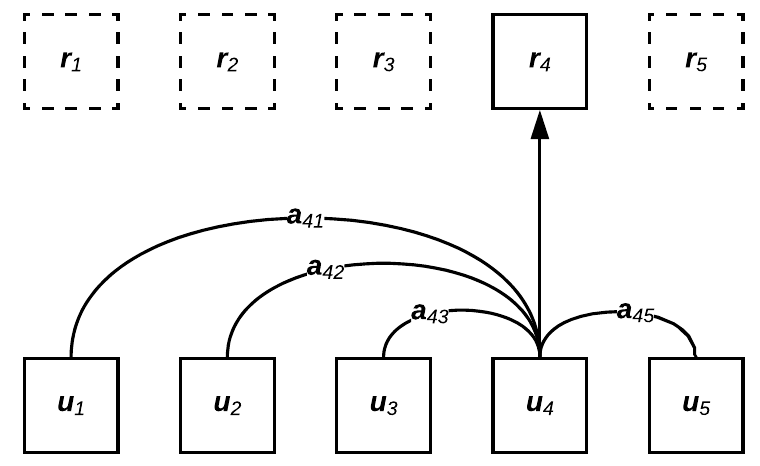}
  \caption{Intra-sentential attention mechanism; $\bm{a}_{ij}$~denotes the
    normalized attention score between tokens $\bm{u}_i$ and $\bm{u}_j$.}
  \label{fig:stratt}
\vspace{-2ex}
\end{figure}

Figure~\ref{fig:stratt} provides a schematic view of the
intra-sentential attention mechanism.  Given a sentence represented as
a sequence of~$n$ word vectors $[\bm{u}_1,\bm{u}_2,\cdots,\bm{u}_n]$,
for each word pair $\langle \bm{u}_i, \bm{u}_j\rangle$, the attention
score $\bm{a}_{ij}$ is estimated as:
\begin{gather}
\bm{f}_{ij} = F(\bm{u}_i, \bm{u}_j)\\
\bm{a}_{ij} = \frac{exp(\bm{f}_{ij})}{\sum_{k=1}^n{exp(\bm{f}_{ik})}} \label{eq:probability}
\end{gather}
where~$F()$ is a function for computing the unnormalized score
$\bm{f}_{ij}$ which is then normalized by calculating a probability
distribution $\bm{a}_{ij}$.  Individual words collect information from their context
based on  $\bm{a}_{ij}$ and obtain a context representation:
\begin{gather}
\bm{r}_{i} = \sum_{j=1}^n\bm{a}_{ij}\bm{u}_j
\end{gather}
where attention score~$\bm{a}_{ij}$ indicates the (dependency) relation
between the~$i$-th and the \mbox{$j$-th-words} and how information
from~$\bm{u}_j$ should be fed into~$\bm{u}_i$.

Despite successful application of the above attention mechanism in
sentiment analysis~\citep{cheng2016long} and entailment
recognition~\citep{parikh2016decomposable}, the structural information
under consideration is shallow, limited to word-word dependencies.
Since attention is computed as a simple probability distribution, it
cannot capture more elaborate structural dependencies such as trees
(or graphs).  \cite{kim2017structured} induce richer internal
structure by imposing structural constraints on the probability
distribution computed by the attention mechanism. Specifically, they
normalize~$\bm{f}_{ij}$ with a projective dependency tree using the
inside-outside algorithm \citep{baker1979trainable}:
\begin{gather}
\bm{f}_{ij} = F(\bm{u}_i, \bm{u}_j)\\
\bm{a} = inside\text{-}outside(\bm{f})\\
\bm{r}_{i} = \sum_{j=1}^n\bm{a}_{ij}\bm{u}_j
\end{gather}
This process is differentiable, so the model can be trained end-to-end
and learn structural information without relying on a parser. However,
efficiency is a major issue, since the inside-outside algorithm has
time complexity~$O(n^3)$ (where $n$~represents the number of tokens)
and does not lend itself to easy parallelization. The high order
complexity renders the approach impractical for real-world
applications.

\section{Encoding Text Representations}

In this section we present our document representation model. We
follow previous work \citep{tang2015document,yang2016hierarchical} in
modeling documents \emph{hierarchically} by first obtaining
representations for sentences and then composing those into a document
representation. Structural information is taken into account while
learning representations for both sentences and documents and an
attention mechanism is applied on both words within a sentence and
sentences within a document. The general idea is to force pair-wise
attention between text units to form a non-projective dependency tree,
and automatically induce this tree for different natural language
processing tasks in a differentiable way. In the following, we first
describe how the attention mechanism is applied to sentences, and then
move on to present our document-level model.




\subsection{Sentence Model}
Let~$T=[\bm{u}_1, \bm{u}_2, \cdots, \bm{u}_n]$ denote a sentence
containing a sequence of words, each represented by a vector~$\bm{u}$,
which can be pre-trained on a large corpus.  Long Short-Term Memory
Neural Networks \citep[LSTMs;][]{hochreiter1997long} have been
successfully applied to various sequence modeling tasks ranging from
machine translation \citep{bahdanau2014neural}, to speech
recognition~\citep{graves2013speech}, and image caption
generation~\citep{xu2015show}.  In this paper we use bidirectional
LSTMs as a way of representing elements in a sequence (i.e., words or
sentences) together with their contexts, capturing the element and an
``infinite'' window around it. Specifically, we run a bidirectional
LSTM over sentence~$T$, and take the output vectors
\mbox{$[\bm{h}_1, \bm{h}_2, \cdots, \bm{h}_n]$} as the representations
of words in~$T$, where $\bm{h}_t \in \mathbb{R}^k$ is the output
vector for word $\bm{u}_t$ based on its context.




We then exploit the structure of~$T$ which we induce based on an
attention mechanism detailed below to obtain more precise
representations.  Inspired by recent
work~\citep{daniluk2017frustratingly,miller2016key}, which shows that
the conventional way of using LSTM output vectors for calculating both
attention and encoding word semantics is overloaded and likely to
cause performance deficiencies, we decompose the LSTM output vector in
two parts:
\begin{gather}
\label{eq:separate}
[\bm{e}_t, \bm{d}_t] = \bm{h}_t
\end{gather}
where $\bm{e}_t \in \mathbb{R}^{k_t}$, the semantic vector, encodes
semantic information for specific tasks, and $\bm{d}_t\in
\mathbb{R}^{k_s}$, the structure vector, is used to calculate structured
attention.

\begin{figure}[t]
  \centering
  \includegraphics[width=3.1in]{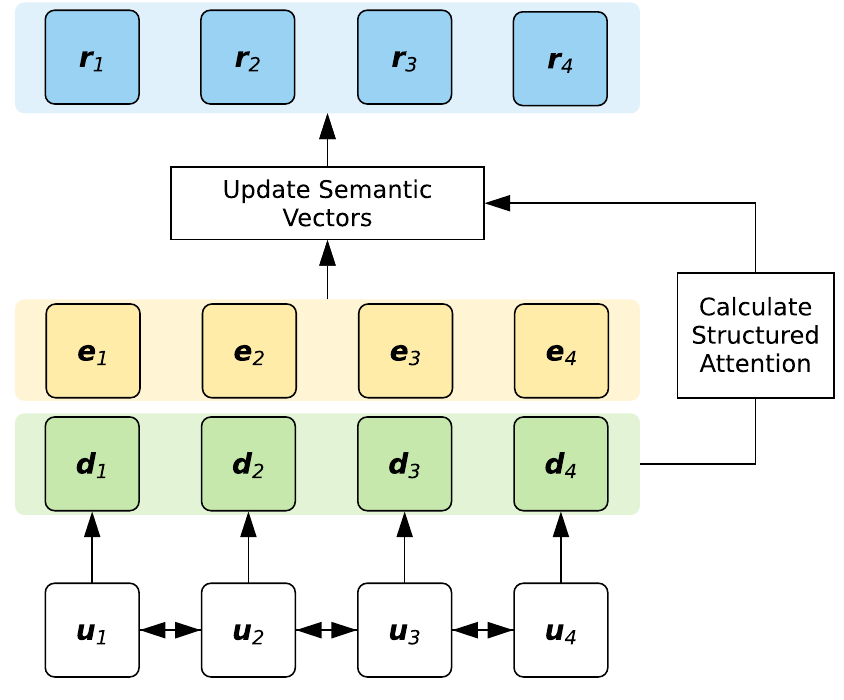}
  \caption{Sentence representation model: $\bm{u}_t$ is the input
    vector for the $t$-th word, $\bm{e}_t$ and $\bm{d}_t$ are semantic
    and structure vectors, respectively. }
  \label{fig:overview}
\vspace{-2ex}
\end{figure}


We use a series of operations based on the Matrix-Tree Theorem
\citep{tutte1984graph} to incorporate the structural bias of
non-projective dependency trees into the attention weights.  We
constrain the probability distributions~$\bm{a}_{ij}$ (see
Equation~\eqref{eq:probability}) to be the posterior marginals of a
dependency tree structure. We then use the normalized structured
attention, to build a context vector for updating the semantic vector
of each word, obtaining new representations $[\bm{r}_1, \bm{r}_2,
\cdots, \bm{r}_n]$.  An overview of the model is presented in
Figure~\ref{fig:overview}. We describe the attention mechanism in
detail in the following section. 


\subsection{Structured Attention Mechanism}

Dependency representations of natural language are a simple yet
flexible mechanism for encoding words and their syntactic relations
through directed graphs. Much work in descriptive linguistics
\citep{Tesniere:1959,Melcuk:1988} has advocated their suitability for
representing syntactic structure across languages. A primary advantage
of dependency representations is that they have a natural mechanism
for representing discontinuous constructions arising from long
distance dependencies or free word order through non-projective
dependency edges.



More formally, building a dependency tree amounts to finding latent
variables~$z_{ij}$ for all $i\neq j$, where word~$i$ is the parent
node of word~$j$, under some global constraints, amongst which the
\mbox{single-head} constraint is the most important, since it forces
the structure to be a rooted tree. We use a variant of Kirchhoff's
Matrix-Tree Theorem~\citep{tutte1984graph,koo2007structured} to
calculate the marginal probability of each dependency edge
$P(z_{ij}=1)$ of a non-projective dependency tree, and this
probability is used as the attention weight that decides how much
information is collected from child unit~$j$ to the parent unit~$i$.



We first calculate unnormalized attention scores~$\bm{f}_{ij}$ with
structure vector~$\bm{d}$ (see Equation~\eqref{eq:separate}) via a
bilinear function:
\begin{gather}
\bm{t}_p = tanh(\bm{W}_p\bm{d}_{i})\\
\bm{t}_c = tanh(\bm{W}_c\bm{d}_{j})\\
\bm{f}_{ij} = \bm{t}_p^T\bm{W}_a\bm{t}_c
\end{gather}
where $\bm{W}_p \in \mathbb{R}^{{k_s}*{k_s}}$ and $\bm{W}_c \in
\mathbb{R}^{{k_s}*{k_s}}$ are the weights for building the
representation of parent and child nodes.  $\bm{W}_a \in
\mathbb{R}^{{k_s}*{k_s}}$ is the weight for the bilinear
transformation.  $\bm{f} \in \mathbb{R}^{n*n}$ can be viewed as a weighted adjacency matrix
for a graph~$G$ with $n$ nodes where each node corresponds to a word in a sentence.  We
also calculate the root score $\bm{f}^r_i$, indicating the
unnormalized possibility of a node being the root:
\begin{flalign}
\bm{f}^r_i = \bm{W}_r\bm{d}_{i}
\end{flalign}
where $\bm{W}_r \in \mathbb{R}^{1*{k_s}}$. We calculate~$P(z_{ij}=1)$,
the marginal probability of the dependency edge, following
\cite{koo2007structured}:
\begin{align}
\bm{A}_{ij}&=\begin{cases}
0 & \text{if }i=j\\
exp(\bm{f}_{ij}) & \text{otherwise}
\end{cases}\\
\bm{L}_{ij}&=\begin{cases}
  \sum_{i'=1}^n \bm{A}_{i'j} & \text{if }i=j \\
  - \bm{A}_{ij} & \text{otherwise}
\end{cases}\\
\bar{\bm{L}}_{ij}&=\begin{cases}
  exp(\bm{f}^r_i) & i=1\\
  \bm{L}_{ij} & i>1
\end{cases}\\
\nonumber P(z&_{ij}=1) = (1-\delta_{1,j})\bm{A}_{ij}[\bar{\bm{L}}^{-1}]_{jj}\\
 &\quad \quad \quad \quad -(1-\delta_{i,1})\bm{A}_{ij}[\bar{\bm{L}}^{-1}]_{ji}\\
 \nonumber P(r&oot(i)) = exp(f^i_r)[\bar{\bm{L}}^{-1}]_{i1}
 \end{align}
 where $1\leq i\leq n, 1\leq j\leq n$. $\bm{L} \in \mathbb{R}^{n*n}$ is the Laplacian matrix for
 graph~$G$ and $\bar{\bm{L}} \in \mathbb{R}^{n*n}$~is a variant of $\bm{L}$ that takes the
 root node into consideration, and $\delta$~is the Kronecker delta.
 The key for the calculation to hold is for~$\bm{L}^{ii}$, the minor
 of the Laplacian matrix~$\bm{L}$ with respect to row $i$ and column
 $i$, to be equal to the sum of the weights of all directed spanning
 trees of~$G$ which are rooted at~$i$.  $P(z_{ij}=1)$ is the marginal
 probability of the dependency edge between the $i$-th and
 $j$-th~words. $P(root(i)=1)$ is the marginal probability of the
 $i$-th word headed by the root of the tree.  Details of the proof can
 be found in \cite{koo2007structured}.

 We denote the marginal probabilities~$P(z_{ij}=1)$ as $\bm{a}_{ij}$
 and $P(root(i))$ as $\bm{a}^r_{i}$. This can be interpreted as
 attention scores which are constrained to converge to a structured
 object, a non-projective dependency tree, in our case.  We update the
 semantic vector $\bm{e}_i$ of each word with structured attention:
\begin{align}
\bm{p}_i &= \sum_{k=1}^n\bm{a}_{ki}\bm{e}_k+\bm{a}^r_{i}\bm{e}_{root}\\
\bm{c}_i &= \sum_{k=1}^n\bm{a}_{ik}\bm{e}_i\\
\bm{r}_i &= tanh(\bm{W}_r[\bm{e}_i, \bm{p}_i, \bm{c}_i])
\end{align}
where $\bm{p}_i \in \mathbb{R}^{k_e}$ is the context vector gathered
from possible parents of $u_i$ and $\bm{c}_i \in \mathbb{R}^{k_e}$ the
context vector gathered from possible children, and $\bm{e}_{root}$ is
a special embedding for the root node. The context vectors are
concatenated with $\bm{e}_i$ and transformed with weights
$\bm{W}_r \in \mathbb{R}^{{k_e}*3{k_e}}$ to obtain the updated
semantic vector $\bm{r}_i \in \mathbb{R}^{k_e}$ with rich structural
information (see Figure~\ref{fig:overview}).

\subsection{Document Model}

We build document representations hierarchically: sentences are
composed of words and documents are composed of sentences. Composition
on the document level also makes use of structured attention in the
form of a dependency graph. Dependency-based representations have been
previously used for developing discourse parsers
\citep{li-EtAl:2014:P14-11,hayashi2016empirical}
and in applications such as summarization
\citep{hirao-EtAl:2013:EMNLP}.

\begin{figure}[t]
  \centering
  \includegraphics[width=2.1in]{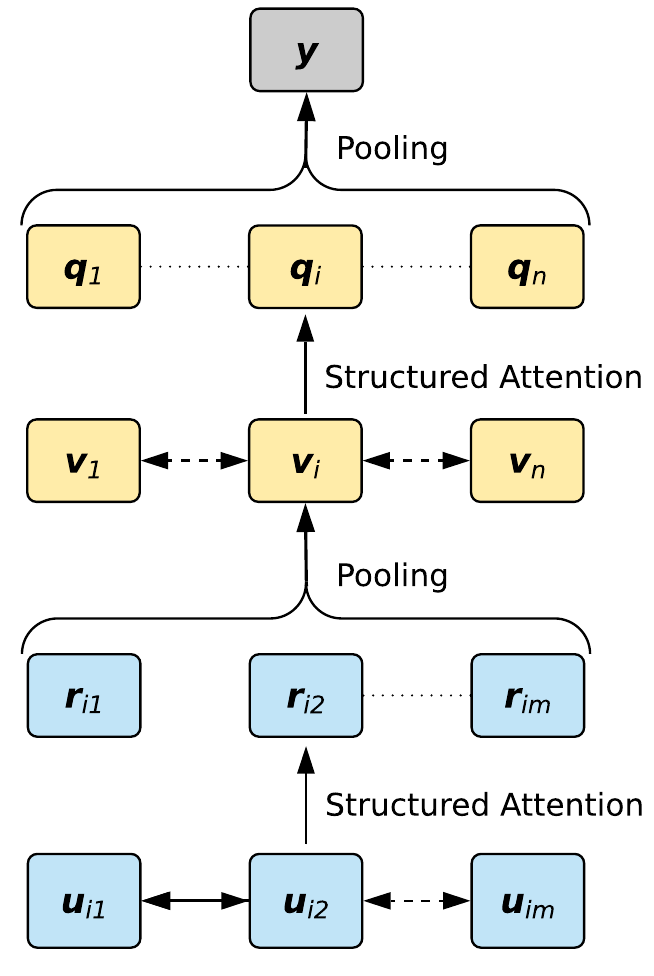} 
  \caption{Document representation model.}
  \label{fig:doc}
\vspace{-2ex}
\end{figure}

As illustrated in Figure~\ref{fig:doc}, given a document with $n$
sentences $[s_{1},s_{2},\cdots,s_{n}]$, for each sentence~$s_i$, the
input is a sequence of word embeddings
$[\bm{u}_{i1},\bm{u}_{i2},\cdots,\bm{u}_{i{m}}]$, where~$m$ is the
number of tokens in~$s_i$.  By feeding the embeddings into a
sentence-level bi-LSTM and applying the proposed structured attention
mechanism, we obtain the updated semantic vector
$[\bm{r}_{i1},\bm{r}_{i2},\cdots,\bm{r}_{i{m}}]$.  Then a pooling
operation produces a fixed-length vector $\bm{v}_i$ for each sentence.
Analogously, we view the document as a sequence of sentence vectors
$[\bm{v}_1, \bm{v}_2,\cdots, \bm{v}_n]$ whose embeddings are fed to a
document-level bi-LSTM. Application of the structured attention
mechanism creates new semantic vectors
$[\bm{q}_{1},\bm{q}_{2},\cdots,\bm{q}_{n}]$ and another pooling operation
yields the final document representation~$\bm{y}$.


\subsection{End-to-End Training}
Our model can be trained in an end-to-end fashion since all operations
required for computing structured attention and using it to update the
semantic vectors are differentiable.  In contrast to in
\cite{kim2017structured}, training can be done efficiently. The major
complexity of our model lies in the computation of the gradients of
the the inverse matrix. Let $A$~denote a matrix depending on a real
parameter~$x$; assuming all component functions in~$A$ are
differentiable, and $A$~is invertible for all possible values, the
gradient of~$A$ with respect respect to~$x$ is:
\begin{gather}
\frac{dA^{-1}}{dx} = -A^{-1}\frac{dA}{dx}A^{-1}
\end{gather}
Multiplication of the three matrices and matrix inversion can be
computed efficiently on modern parallel hardware architectures such as
GPUs. In our experiments, computation of structured attention takes
only~1/10 of training time.


\section{Experiments}

In this section we present our experiments for evaluating the
performance of our model. Since sentence representations constitute
the basic building blocks of our document model, we first evaluate the
performance of structured attention on a sentence-level task, namely
natural language inference. We then assess the document-level
representations obtained by our model on a variety of classification
tasks representing documents of different length, subject matter, and
language. Our code is available at \url{https://github.com/nlpyang/structured}.

\subsection{Natural Language Inference}
The ability to reason about the semantic relationship between two
sentences is an integral part of text understanding. We therefore
evaluate our model on recognizing textual entailment, i.e., whether
two premise-hypothesis pairs are entailing, contradictory, or neutral.
For this task we used the Stanford Natural Language Inference (SNLI)
dataset~\citep{bowman-EtAl:2015:EMNLP}, which contains
premise-hypothesis pairs and target labels indicating their relation.
After removing sentences with unknown labels, we obtained~549,367 pairs
for training, 9,842~for development and 9,824 for testing.

\begin{table*}[t]
\centering
\begin{tabular}{|@{~}llc@{~}|}
  \hline
  Models                                 & Acc  & $\theta$\\ \hline\hline
  Classifier with handcrafted features \citep{bowman-EtAl:2015:EMNLP}
  & 78.2 & --- \\
  300D LSTM encoders \citep{bowman-EtAl:2015:EMNLP}          & 80.6  &
  3.0M \\
  300D  Stack-Augmented Parser-Interpreter Neural Net 
  \citep{bowman-EtAl:2016:P16-1} & 83.2  & 3.7M \\
  100D LSTM with inter-attention \citep{rocktaschel2015reasoning} &
  83.5 & 252K \\
  200D Matching LSTMs \citep{wang2015learning} & 86.1   & 1.9M  \\
  450D LSTMN with deep attention fusion \citep{cheng2016long} & 86.3 &
  3.4M \\
  Decomposable Attention over word embeddings
  \citep{parikh2016decomposable}            & 86.8  & 582K    \\
  Enhanced BiLSTM Inference Model \citep{chen2016enhancing}       &
  \textbf{88.0}  &   4.3M\\ \hline \hline
  175D No Attention                      & 85.3   & 600K \\
  175D Simple intra-sentence attention                      & 86.2  & 1.1M \\
  100D Structured intra-sentence attention with Inside-Outside & 86.8 &  1.2M \\
  175D Structured intra-sentence attention with Matrix Inversion  &
  \textbf{86.9}&   1.1M  \\ \hline
\end{tabular}
\caption{Test accuracy on the SNLI dataset  and number of parameters~$\theta$
  (excluding embeddings). Wherever available we also provide the size
  of the recurrent unit.}
\label{exp2}
\end{table*}

Sentence-level representations obtained by our model (with structured
attention) were used to encode the premise and hypothesis by modifying
the model of~\protect\cite{parikh2016decomposable} as follows.  Let
$[\bm{x}^p_1, \cdots, \bm{x}^p_n]$ and $[\bm{x}^h_1, \cdots,
\bm{x}^h_m]$ be the input vectors for the premise and hypothesis,
respectively. Application of structured attention yields new vector
representations $[\bm{r}^p_1, \cdots, \bm{r}^p_n]$ and $[\bm{r}^h_1,
\cdots, \bm{r}^h_m]$. Then we combine these two vectors with
inter-sentential attention, and apply an average pooling operation:
\begin{gather}
\bm{o}_{ij} = MLP(\bm{r}^p_i)^TMLP(\bm{r}^h_j)\\
\bar{\bm{r}}^p_i = [\bm{r}^p_i, \sum_{j=1}^m\frac{exp(\bm{o}_{ij})}{\sum_{k=1}^mexp(\bm{o}_{ik})}]\\
\bar{\bm{r}}^h_i = [\bm{r}^h_i, \sum_{i=1}^m\frac{exp(\bm{o}_{ij})}{\sum_{k=1}^mexp(\bm{o}_{kj})}]\\
\bm{r}^p = \sum_{i=1}^ng(\bar{\bm{r}}^p_i),\quad\bm{r}^h = \sum_{i=1}^mg(\bar{\bm{r}}^h_i)
\end{gather}
where $MLP()$ is a two-layer perceptron with a~$ReLU$ activation
function. The new representations $\bm{r}^p, \bm{r}^h$ are then
concatenated and fed into another two-layer perceptron with a softmax
layer to obtain the predicted distribution over the labels.

\begin{table}[t]
\centering
\begin{tabular}{|l|ll|}
\hline
\multirow{2}{*}{Models} & \multicolumn{2}{c|}{Speed}                                                   \\ \cline{2-3} 
                         & \multicolumn{1}{c}{Max} & \multicolumn{1}{c|}{Avg} \\ \hline
No Attention         &                           0.0050                       &      0.0033                    \\

Simple  Attention       &                    0.0057                              &0.0042                          \\
Matrix Inversion         &  0.0070                                           &        0.0045                  \\ 
Inside-Outside         &      0.1200                                            &     0.0380                     \\\hline
\end{tabular}
\caption{Comparison of speed of different models on the SNLI
  testset. The unit of measurement is seconds per instance. All
  results were obtained on a GeForce GTX  TITAN X (Pascal) GPU.}  
\label{speed}
\vspace{-2ex}
\end{table}

The hidden size of the LSTM was set to 150. The dimensions of the
semantic vector were~100 and the dimensions of structure vector
were~50.  We used pretrained 300-D Glove 840B
\citep{pennington2014glove} vectors to initialize the word
embeddings. All parameters (including word embeddings) were updated
with Adagrad~\citep{duchi2011adaptive}, and the learning rate was set
to~0.05. The hidden size of the two-layer perceptron was set to~200
and dropout was used with ratio~0.2.  The mini-batch size was~32.

We compared our model (and variants thereof) against several related
systems. Results (in terms of 3-class accuracy) are shown in
Table~\ref{exp2}. Most previous systems employ LSTMs and do not
incorporate a structured attention component. Exceptions include
\cite{cheng2016long} and \cite{parikh2016decomposable} whose models
include intra-attention encoding relationships between words within
each sentence (see Equation~\eqref{eq:probability}). It is also worth
noting that some models take structural information into account in
the form of parse trees
\citep{bowman-EtAl:2016:P16-1,chen2016enhancing}.
The second block of Table~\ref{exp2} presents a version of our model
without an intra-sentential attention mechanism as well as three
variants with attention, assuming the structure of word-to-word
relations and dependency trees. In the latter case we compare our
matrix inversion based model against Kim et al.'s
(\citeyear{kim2017structured}) inside-outside attention
model. Consistent with previous work
\citep{parikh2016decomposable,cheng2016long}, we observe that simple
attention brings performance improvements over no
attention. Structured attention further enhances performance. Our own
model with tree matrix inversion slightly outperforms the
inside-outside model of \cite{kim2017structured}, overall achieving
results in the same ballpark with related LSTM-based models
\citep{parikh2016decomposable,cheng2016long,chen2016enhancing}.


Table~\ref{speed} compares the running speed of the models shown in
the second block of Table~\ref{exp2}.  As can be seen matrix inversion
does not increase running speed over the simpler attention mechanism
and is considerably faster compared to inside-outside. The latter is
10--20 times slower than our model on the same platform.


\begin{table}[t]
\centering
\begin{tabular}{|l|rrrr|}
\hline
Dataset    & \multicolumn{1}{c}{\#class} & \multicolumn{1}{c}{\#docs} & \multicolumn{1}{c}{\#s/d} & \multicolumn{1}{c|}{\#w/d} \\ \hline
Yelp        & 5       & 335K      & 8.9         & 151.6       \\
IMDB        & 10      & 348K      & 14.0        & 325.6       \\
CZ Movies   & 3       & 92K       & 3.5         & 51.2        \\
Debates     & 2       & 1.6K      & 22.7        & 519.2       \\ \hline
\end{tabular}
\caption{Dataset statistics; \#class is the number of classes per
  dataset,  \#docs denotes the number of documents;  \#s/d  and \#w/d
  represent the average number of sentences and  words per document.}
\label{stat}
\vspace{-2ex}
\end{table}

\begin{table*}[t]
\centering
\begin{tabular}{|@{~}l@{~}|@{~}c@{~}|c@{~}|@{~}c@{~}|@{~}c@{~}|@{~}c@{~}|}
  \hline
  Models                         & Yelp &IMDB &CZ  Movies& Debates&        $\theta$\\\hline
  Feature-based classifiers & 59.8 &40.9 & 78.5&74.0&  ---\\
  Paragraph vector \citep{tang2015document} & 57.7&   34.1 & ---&----&   ---\\
  Convolutional neural network   \citep{tang2015document}     & 59.7&     ---&---&---&         --- \\
  Convolutional gated RNN \citep{tang2015document} & 63.7&     42.5& ---& ---&---\\
  LSTM gated RNN \citep{tang2015document}  & 65.1&     45.3&---&---&       ---\\
  RST-based recursive neural network \citep{ji2017neural}&---&---&---&75.7&---\\
  75D Hierarchical attention networks \citep{yang2016hierarchical}& 68.2
  &   \textbf{49.4}&80.8&74.0 &273K\\\hline \hline
  75D No Attention   & 66.7   & 47.5& 80.5&73.7&  330K\\
  100D Simple Attention   & 67.7   & 48.2 &81.4&75.3 & 860K\\ 
  100D Structured Attention (sentence-level)& 68.0  &48.8 &81.5&74.6 & 842K\\ 
  100D Structured Attention (document-level) & 67.8  &48.6 &81.1&75.2 &842K\\ 
  100D Structured Attention (both levels) &\textbf{68.6}& 49.2 &\textbf{82.1}&\textbf{76.5} & 860K\\ \hline
\end{tabular}
\caption{Test accuracy on four datasets and number of parameters
  $\theta$ (excluding embeddings). Regarding feature-based
  classification methods, results on Yelp and IMDB are taken from
  \cite{tang2015document}, on CZ movies from
  \cite{brychcin2013unsupervised},  and Debates from
  \cite{yogatama-smith:2014:P14-1}. Wherever available we also provide the size
  of the recurrent unit (LSTM or GRU).}
\label{exp3}
\vspace{-2ex}
\end{table*}

\subsection{Document Classification}
In this section, we evaluate our document-level model on a variety of
classification tasks. We selected four datasets which we describe
below. Table~\ref{stat} summarizes some statistics for each dataset.

\paragraph{Yelp reviews} were obtained from the 2013 Yelp Dataset
Challenge.  This dataset contains restaurant reviews, each associated
with human ratings on a scale from~1 (negative) to 5~(positive) which
we used as gold labels for sentiment classification; we followed the
preprocessing introduced in~\cite{tang2015document} and report
experiments on their training, development, and testing partitions
(80/10/10).

\paragraph{IMDB reviews} were obtained from~\cite{diao2014jointly},
who randomly crawled  reviews for~50K movies. Each review is
associated with user ratings ranging from~1 to~10. 

\paragraph{Czech reviews} were obtained
from~\cite{brychcin2013unsupervised}. The dataset contains reviews
from the Czech Movie Database\footnote{\url{http://www.csfd.cz/}} each
labeled as positive, neutral, or negative. We include Czech in our
experiments since it has more flexible word order compared to English,
with non-projective dependency structures being more frequent.
Experiments on this dataset perform 10-fold cross-validation following
previous work~\citep{brychcin2013unsupervised}.

\paragraph{Congressional floor debates} were obtained from a corpus
originally created by~\cite{thomas2006get} which contains transcripts
of U.S. floor debates in the House of Representatives for the year
2005. Each debate consists of a series of speech segments, each
labeled by the vote (``yea'' or ``nay'') cast for the proposed bill by
the the speaker of each segment. We used the pre-processed corpus
from~\cite{yogatama-smith:2014:P14-1}.\footnote{\url{http://www.cs.cornell.edu/~ainur/data.html}}

\begin{figure*}[t]
  \centering
  \includegraphics[width=6in]{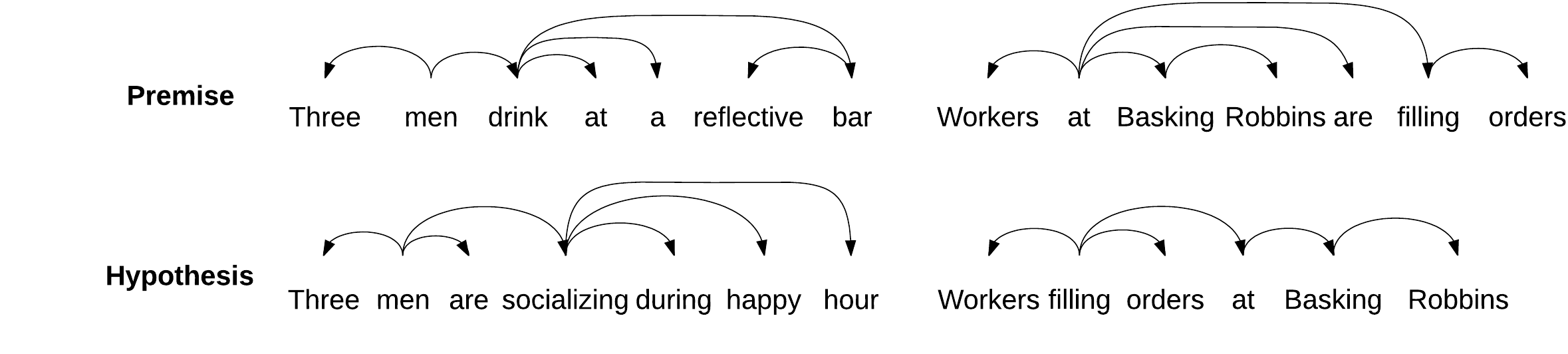} 
  \caption{Dependency trees induced by our model on the SNLI test set.}
  \label{fig:ana1}
\vspace{-2ex}
\end{figure*}


Following previous work~\citep{yang2016hierarchical}, we only retained
words appearing more than five times in building the vocabulary and
replaced words with lesser frequencies with a special UNK token. Word
embeddings were initialized by training
word2vec~\citep{mikolov2013distributed} on the training and validation
splits of each dataset.  In our experiments, we set the word embedding
dimension to be~200 and the hidden size for the sentence-level and
document-level LSTMs to~100 (the dimensions of the semantic and
structure vectors were set to~75 and~25, respectively).  We used a
mini-batch size of~32 during training and documents of similar length
were grouped in one batch.  Parameters were optimized with Adagrad
\citep{duchi2011adaptive}, the learning rate was set to~0.05.  We
used~$L_2$ regularization for all parameters except word embeddings
with regularization constant set to $1e^{-4}$. Dropout was applied on
the input and output layers with dropout rate~0.3.

Our results are summarized in Table~\ref{exp3}.  We compared our model
against several related models covering a wide spectrum of
representations including word-based ones (e.g., paragraph vector and
CNN models) as well as hierarchically composed ones (e.g., a CNN or
LSTM provides a sentence vector and then a recurrent neural network
combines the sentence vectors to form a document level representation
for classification).  Previous state-of-the-art results on the three
review datasets were achieved by the hierarchical attention network of
\cite{yang2016hierarchical}, which models the document hierarchically
with two GRUs and uses an attention mechanism to weigh the importance
of each word and sentence.  On the debates corpus, \cite{ji2017neural}
obtained best results with a recursive neural network model operating
on the output of an RST parser. Table~\ref{exp3} presents three
variants\footnote{We do not report comparisons with the
  inside-outside approach on document classification tasks due to
  its prohibitive computation cost leading to 5 hours of training for
  one epoch.} of our model, one with structured attention on the
sentence level, another one with structured attention on the document
level and a third model which employs attention on both levels. As can
be seen, the combination is beneficial achieving best results on three
out of four datasets. Furthermore, structured attention is superior to
the simpler word-to-word attention mechanism, and both types of
attention bring improvements over no attention.   The structured
attention approach is also very efficient, taking only 20 minutes for
one training epoch on the largest dataset.


\subsection{Analysis of Induced Structures}
\label{sec:analys-induc-struct}

To gain further insight on structured attention, we inspected the
dependency trees it produces. Specifically, we used the
Chu-Liu-Edmonds algorithm~\citep{chu1965shortest,edmonds1967optimum}
to extract the maximum spanning tree from the attention scores. We
report various statistics on the characteristics of the induced trees
across different tasks and datasets. We also provide examples of tree
output, in an attempt to explain how our model uses dependency
structures to model text.


\begin{table}[t]
\centering
\begin{tabular}{|ll|cr|}
\hline
                                            &         & Parser                   & Attention\\ \hline
\multicolumn{2}{|l|}{Projective}                       & \multicolumn{1}{c}{---} & 51.4\%      \\ 
\multicolumn{2}{|l|}{Height}                           & 8.99      & 5.78\hspace*{.13cm}      \\ \hline
\multicolumn{1}{|l|}{\multirow{6}{*}{Nodes}} & depth 1 & 9.8\%                 & 8.4\%      \\
\multicolumn{1}{|l|}{}                       & depth 2 & 15.0\%                  & 19.7\%     \\
\multicolumn{1}{|l|}{}                       & depth 3 & 12.8\%                & 22.4\%     \\
\multicolumn{1}{|l|}{}                       & depth 4 & 12.5\%                & 23.4\%     \\
\multicolumn{1}{|l|}{}                       & depth 5 & 12.0\%                  & 14.4\%     \\
\multicolumn{1}{|l|}{}                       & depth 6 & 10.3\%                & 4.5\%      \\\hline
\multicolumn{2}{|l|}{Same Edges}                       & \multicolumn{2}{|c|}{38.7\%}      \\  \hline
\end{tabular}
\caption{Descriptive statistics for dependency trees produced by our
  model and the Stanford parser \citep{manning2014stanford} on  the SNLI test set.}
\label{ana1}
\vspace{-2ex}
\end{table}

\paragraph{Sentence Trees} We compared the dependency trees obtained
from our model with those produced by a state-of-the-art dependency
parser trained on the English Penn Treebank. Table~\ref{ana1} presents
various statistics on the depth of the trees produced by our model on
the SNLI test set and the Stanford dependency parser
\citep{manning2014stanford}. As can be seen, the induced dependency
structures are simpler compared to those obtained from the Stanford
parser. The trees are generally less deep (their height is 5.78
compared to 8.99 for the Stanford parser), with the majority being of
depth 2--4.  Almost half of the induced trees have a projective
structure, although there is nothing in the model to enforce this
constraint. We also calculated the percentage of head-dependency edges
that are identical between the two sets of trees. Although our model
is not exposed to annotated trees during training, a large number of
edges agree with the output of the Stanford parser.

\begin{table}[t]
\centering
\begin{tabular}{|@{}l@{~}l|c@{~}c@{~}c@{~}c@{}|}
\hline
                      &         & Yelp   & IMDB  &  CZ Movies & Debates \\ \hline
\multicolumn{2}{|@{}l@{}|}{Projective}                    & 79.6\% & 74.9\% &82.8\%& 62.4\%  \\ 
\multicolumn{2}{|@{}l@{}|}{Height}                         & 2.81   & 3.34   &1.50& 3.58    \\ \hline
\multicolumn{1}{|@{}l|}{\multirow{4}{*}{Nodes}} & depth 2 & 15.1\% & 13.6\% &25.7\%& 12.8\%  \\
\multicolumn{1}{|@{}l@{}|}{}    & depth 3 & 55.6\% & 46.8\% &57.1\%& 30.2\%  \\
\multicolumn{1}{|@{}l@{}|}{} & depth 4 & 22.3\% & 32.5\% &11.3\%& 40.8\%  \\
\multicolumn{1}{|@{}l@{}|}{}  & depth 5 & 3.2\%  & 4.1\%  &5.8\%& 14.8\%  \\ \hline
\end{tabular}
\caption{Descriptive statistics for induced document-level dependency trees
  across datasets.} 
\label{ana2}
\vspace{-2ex}
\end{table}

Figure~\ref{fig:ana1} shows examples of dependency trees induced on
the SNLI dataset.  Although the model is trained without ever being
exposed to a parse tree, it is able to learn  plausible dependency
structures via the attention mechanism.  Overall we observe that the
induced trees differ from linguistically motivated ones in the types
of dependencies they create which tend to be of shorter length. The
dependencies obtained from structured attention are more direct as
shown in the first premise sentence in Figure~\ref{fig:ana1} where
words \textit{at} and \textit{bar} are directly connected to the verb
\textit{drink}. This is perhaps to be expected since the attention
mechanism uses the dependency structures to collect information from
other words, and the direct links will be more effective.

\begin{figure}[t!]
  \centering
\begin{tabular}{@{}l@{}l@{}}
 (a) & \includegraphics[width=3in]{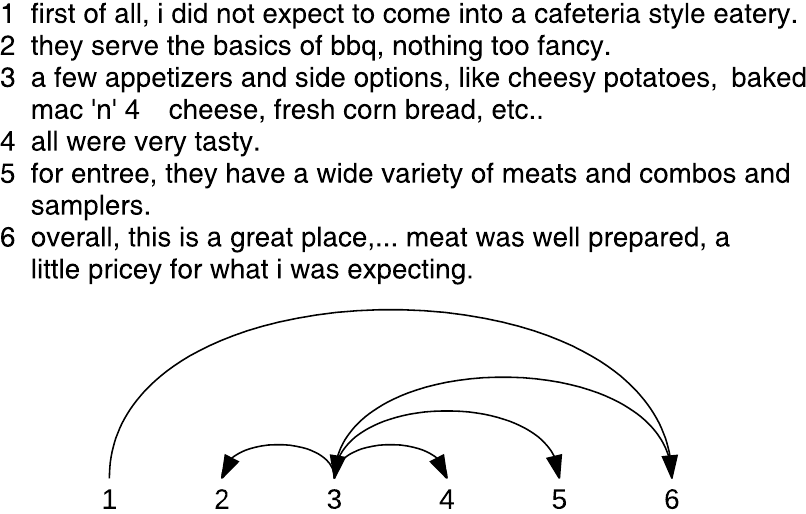}\\\\\vspace{-2ex}
(b)&  \includegraphics[width=3in]{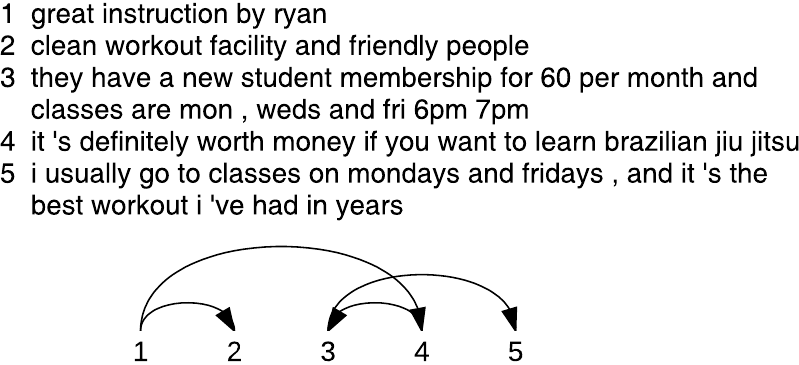} \\\\\vspace{-2ex}
(c)&   \includegraphics[width=3.1in]{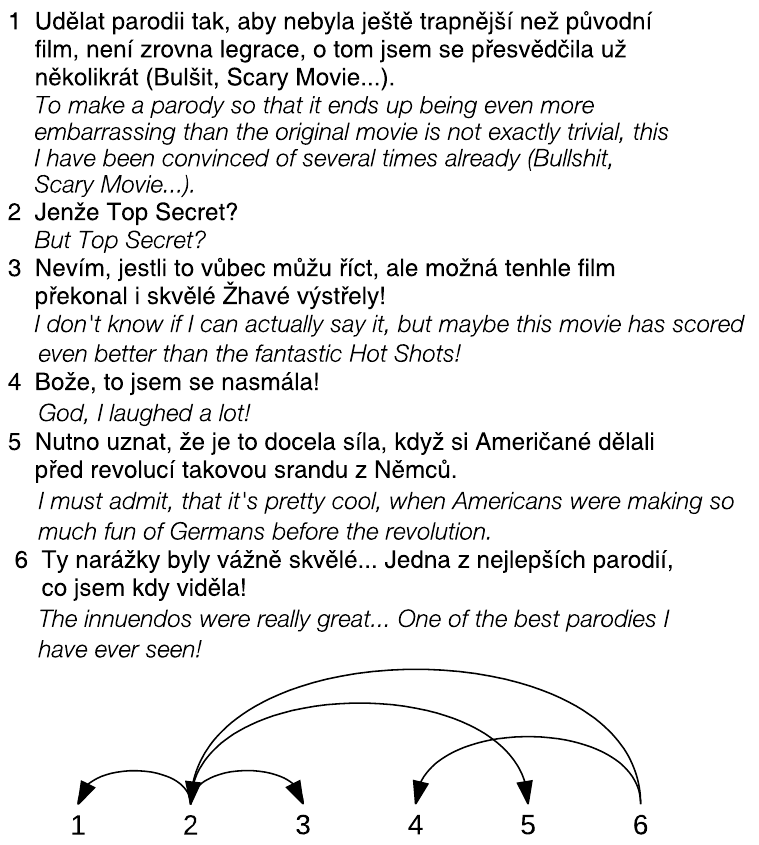}\\
\end{tabular}
\caption{Induced dependency trees for three documents taken from Yelp
  (a,b) and the Czech Movies dataset (c). English translations are in
  italics.}
  \label{fig:ana3}
\vspace{-1ex}
\end{figure}

\paragraph{Document Trees} 

We also used the Chu-Liu-Edmonds algorithms to obtain document-level
dependency trees.  Table~\ref{ana2} summarizes various characteristics
of these trees. For most datasets, document-level trees are not very
deep, they mostly contain up to nodes of depth~3. This is not
surprising as the documents are relatively short (see
Table~\ref{stat}) with the exception of debates which are longer and
the induced trees more complex. The fact that most documents exhibit
simple discourse structures is further corroborated by the large
number (over~70\%) of projective trees induced on Yelp, IMBD, and CZ
Movies datasets. Unfortunately, our trees cannot be directly compared
with the output of a discourse parser which typically involves a
segmentation process splitting sentences into smaller units. Our trees
are constructed over entire sentences, and there is no mechanism
currently in the model to split sentences into discourse units.


Figure~\ref{fig:ana3} shows examples of document-level trees taken
from Yelp and the Czech Movie dataset.  In the first tree, most edges
are examples of the ``elaboration'' discourse relation, i.e., the
child presents additional information about the parent. The second
tree is non-projective, the edges connecting sentences~1 and~4 and
3~and~5 cross.  The third review, perhaps due to its colloquial
nature, is not entirely coherent. However, the model manages to link
sentences 1~and~3 to sentence~2, i.e.,~the movie being discussed; it
also relates sentence 6~to~4, both of which express highly positive
sentiment.



\section{Conclusions}
\label{sec:conclusions}

In this paper we proposed a new model for representing documents while
automatically learning rich structural dependencies. Our model
normalizes intra-attention scores with the marginal probabilities of a
non-projective dependency tree based on a matrix inversion process.
Each operation in this process is differentiable and the model can be
trained efficiently end-to-end, while inducing structural information.
We applied this approach to model documents hierarchically,
incorporating both sentence- and document-level structure.
Experiments on sentence and document modeling tasks show that the
representations learned by our model achieve competitive performance
against strong comparison systems. Analysis of the induced tree
structures revealed that they are meaningful, albeit different from
linguistics ones, without ever exposing the model to linguistic
annotations or an external parser.

Directions for future work are many and varied. Given appropriate
training objectives \citep{TACL972}, it should be possible to induce
linguistically meaningful dependency trees using the proposed
attention mechanism. We also plan to explore how document-level trees
can be usefully employed in summarization, e.g., as a means to
represent or even extract important content.

\paragraph{Acknowledgments}
The authors gratefully acknowledge the support of the European
Research Council (award number 681760). We also thank the anonymous
TACL reviewers and the action editor whose feedback helped improve the present paper, 
members of EdinburghNLP for helpful discussions and suggestions, and
Barbora Skarabela for translating the Czech document for us.

\bibliographystyle{acl_natbib} \bibliography{tacl}

\end{document}